%% file: main.tex
\documentclass[conference]{IEEEtran}
\IEEEoverridecommandlockouts
\usepackage{cite}
\usepackage{amsmath,amssymb,amsfonts}
\usepackage{algorithmicx}
\usepackage{algorithm}
\usepackage{algpseudocode}
\usepackage{subcaption}
\usepackage{graphicx}
\usepackage{multirow}
\usepackage{textcomp}
\usepackage{xcolor}
\usepackage{adjustbox}
\usepackage{array}
\usepackage{rotating}
\usepackage{pgfplots}
\usetikzlibrary{patterns}

\definecolor{Blue1}{RGB}{0, 2, 92}
\definecolor{Blue2}{RGB}{151, 0, 94}
\definecolor{Blue3}{RGB}{232, 60, 63}
\definecolor{Blue4}{RGB}{255, 166, 0}

\definecolor{Green1}{RGB}{0, 92, 78}
\definecolor{Green2}{RGB}{57, 130, 67}
\definecolor{Green3}{RGB}{144, 161, 28}
\definecolor{Green4}{RGB}{255, 174, 0}

\pgfplotscreateplotcyclelist{colorlist1}{
	Blue1, style={fill=Blue1}\\
	Blue2, style={fill=Blue2}\\
	Blue3, style={fill=Blue3}\\
	Blue4, style={fill=Blue4}\\
}

\pgfplotscreateplotcyclelist{colorlist2}{
	Green1, style={fill=Green1}\\
	Green2, style={fill=Green2}\\
	Green3, style={fill=Green3}\\
	Green4, style={fill=Green4}\\
}

\newcommand\alf{autoencoder-based low-rank filter-sharing}

\usepackage{physics}
\usepackage{booktabs}
\usepackage{paralist}
\input{data/misc/happinet_tools.tex}

\DeclareMathOperator{\mse}{\texttt{MSE}}
\DeclareMathOperator{\clip}{\texttt{Clip}}

\def\BibTeX{{\rm B\kern-.05em{\sc i\kern-.025em b}\kern-.08em
    T\kern-.1667em\lower.7ex\hbox{E}\kern-.125emX}}
\begin{document}

\title{ALF: Autoencoder-based Low-rank Filter-sharing for Efficient Convolutional Neural Networks\\
\thanks{\**Corresponding Authors}
}

\input{data/misc/authors.tex}

\maketitle

\begin{abstract}
\input{data/00_abstract.tex}
\end{abstract}


\section{Introduction}
\input{data/10_introduction.tex}

\section{Related Work} \label{related}
\input{data/20_relatedwork.tex}


\section{\alf} \label{alf}
\input{data/40_alf.tex}

\section{Experimental Results}
\input{data/50_experiments.tex}

\section{Conclusion}
\input{data/60_conclusion.tex}
\bibliographystyle{ieeetr}
\bibliography{literatur.bib}

\end{document}

%% file: data/misc/happinet_tools.tex
\usepackage{enumitem,amssymb}
\newlist{todolist}{itemize}{2}
\setlist[todolist]{label=$\square$}
\usepackage{pifont}
\newcommand{\cmark}{\ding{51}}%
\newcommand{\xmark}{\ding{55}}%

\newcommand{\etal}{et al.~}

%% file: data/misc/authors.tex
\author{\IEEEauthorblockN{Alexander Frickenstein\textsuperscript{1\**}, Manoj-Rohit Vemparala\textsuperscript{1\**}, Nael Fasfous\textsuperscript{2\**}, Laura Hauenschild\textsuperscript{2\**},\\
Naveen-Shankar Nagaraja\textsuperscript{1}, Christian Unger\textsuperscript{1}, Walter Stechele\textsuperscript{2}}
\IEEEauthorblockA{\textsuperscript{1}\textit{Autonomous Driving, BMW Group,  Munich, Germany} \\
\textsuperscript{2}\textit{Department of Electrical and Computer Engineering, Technical University of Munich, Munich, Germany}\\
\textsuperscript{1}\{$<\mathtt{firstname}>$.$<\mathtt{lastname}>$\}@bmw.de, \textsuperscript{2}\{$<\mathtt{firstname}>$.$<\mathtt{lastname}>$\}@tum.de}
}

%% file: data/00_abstract.tex
Closing the gap between the hardware requirements of state-of-the-art convolutional neural networks and the limited resources constraining embedded applications is the next big challenge in deep learning research. The computational complexity and memory footprint of such neural networks are typically daunting for deployment in resource constrained environments. Model compression techniques, such as pruning, are emphasized among other optimization methods for solving this problem. Most existing techniques require domain expertise or result in irregular sparse representations, which increase the burden of deploying deep learning applications on embedded hardware accelerators. In this paper, we propose the \alf{} technique (ALF). When applied to various networks, ALF is compared to state-of-the-art pruning methods, demonstrating its efficient compression capabilities on theoretical metrics as well as on an accurate, deterministic hardware-model. In our experiments, ALF showed a reduction of 70\% in network parameters, 61\% in operations and 41\% in execution time, with minimal loss in accuracy.

%% file: data/10_introduction.tex
In recent years, deep learning solutions have gained popularity in several embedded applications ranging from robotics to autonomous driving\cite{binarydad}. This holds especially for computer
vision based algorithms~\cite{googlenet}.
These algorithms are typically demanding in terms of their computational complexity and their memory footprint. Combining these two aspects emphasizes the importance of constraining neural networks in terms of model size and computations for efficient deployment on embedded hardware. The main goal of model compression techniques lies in reducing redundancy, while having the desired optimization target in mind.

Hand-crafted heuristics are applied in the field of model compression, however, they severely reduce the search space and can result in a poor choice of design parameters~\cite{learn_weights}. Particularly for convolutional neural networks (CNNs) with numerous layers, such as ResNet1K \cite{resnet}, hand-crafted optimization is prone to result in a sub-optimal solution. In contrast, a learning-based policy would, for example, leverage reinforcement learning (RL) to automatically explore the CNN's design space~\cite{learn_to_prune}. By evaluating a cost function, the RL-agent learns to distinguish between good and bad decisions, thereby converging to an effective compression strategy for a given CNN. However, the effort of designing a robust cost function and the time spent for model exploration render learning-based compression policies as a complex method, requiring domain expertise. 

In this work, we propose the \alf{} technique (ALF) which generates a dense, compressed CNN during task-specific training. ALF uses the inherent properties of sparse autoencoders to compress data in an unsupervised manner. By introducing an information bottleneck, ALF-blocks are used to extract the most salient features of a convolutional layer during training, in order to dynamically reduce the dimensionality of the layer's tensors. To the best of our knowledge, ALF is the first method where autoencoders are used to prune CNNs. After optimization, the resulting model consists of fewer filters in each layer, and can be efficiently deployed on any embedded hardware due to its structural consistency.
The contributions of this work are summarized as follows:
\begin{compactitem}
\item Approximation of weight filters of convolutional layers using ALF-blocks, consisting of sparse autoencoders.
\item A two player training scheme allows the model to learn the desired task while slimming the neural architecture.
\item Comparative analysis of ALF against learning and rule-based compression methods w.r.t. known metrics, as well as layer-wise analysis on hardware model estimates.
\end{compactitem}
\vspace{-1ex}

%% file: data/20_relatedwork.tex
Efforts from both industry and academia have focused on reducing the redundancy, emerging from training deeper and wider network architectures, with the aim of mitigating the challenges of their deployment on edge devices \cite{resource_aware}. Compression techniques such as quantization, low-rank decomposition and pruning can potentially make CNNs more efficient for the deployment on embedded hardware. Quantization aims to reduce the representation redundancy of model parameters and arithmetic~\cite{lognet, bnn, orthruspe}. Quantization and binarization are orthogonal to this work and can be applied in conjunction with the proposed ALF method.
In the following sections, we classify works which use low-rank decomposition and pruning techniques into rule-based and learning-based compression.
  
\subsection{Rule-based Compression}
Rule-based compression techniques are classified as having static or pseudo-static rules, which are followed when compressing a given CNN. Low-rank decomposition or representation techniques have been used to reduce the number of parameters ($\mathtt{Params}$) by creating separable filters across the spatial dimension or reducing cross-channel redundancy in CNNs~\cite{accelerating,tucker}. Hand-crafted pruning can be implemented based on heuristics to compute the saliency of a neuron. Han \etal\cite{learn_weights} determined the saliency of weights based on their magnitude, exposing the superfluous nature of state-of-the-art neural networks. Pruning individual weights, referred to as irregular pruning, leads to inefficient memory accesses, making it impractical for general purpose computing platforms. Regularity in pruning becomes an important criterion towards accelerator-aware optimization. Frickenstein \etal\cite{dsc} propose structured, kernel-wise magnitude pruning along with a scalable, sparse algorithm. 
He \etal\cite{geometric_mean_filter} prunes redundant filters using a geometric mean heuristic. Although the filter pruning scheme is useful w.r.t. hardware implementations, it is challenging to remove filters as they directly impact the input channels of the subsequent layer. Rule-based compression techniques overly generalize the problem at hand. Different CNNs vary in complexity, structure and target task, making it hard to set such \emph{one-size-fits-all} rules when considering the different compression criteria.

\subsection{Learning-based Compression}

Recent works such as~\cite{learn_to_prune} and \cite{amc} have demonstrated that it is difficult to formalize a rule to prune networks. Instead, they expose the pruning process as an optimization problem to be solved through an RL-agent. Through this process, the RL-agent learns the criteria for pruning, based on a given cost function.

Huang \etal\cite{learn_to_prune} represent a CNN as an environment for an RL-agent. An accuracy term and an efficiency term are combined to formulate a non-differentiable policy to train the agent. The target is to maximize the two contrary objectives. Balancing the terms can eventually vary for different models and tasks. An agent needs to be trained individually for each layer. Moreover, layers with multiple channels may slow down the convergence of the agent, rendering the model exploration as a slow and greedy process.
In the work proposed by He \etal\cite{amc}, an RL-agent prunes channels without fine-tuning at intermediate stages. This cuts down the time needed for exploration. Layer characteristics such as size, stride and operations ($\mathtt{OPs}$), serve the agent as input. These techniques require carefully crafted cost functions for the optimization agent. The formulation of the cost functions is non-trivial, requiring expert knowledge and some iterations of trial-and-error. Furthermore, deciding what the agent considers as the environment can present another variable to the process, making it challenging to test many configurations of the problem. As mentioned earlier, each neural network and target task combination present a new and unique problem, for which this process needs to be repeated.

More recently, Neural Architectural Search (NAS) techniques have been successful in optimizing CNN models at design-time. Combined with Hardware-in-the-Loop (HIL) testing, a much needed synergy between efficient CNN design and the target hardware platform is achieved.
Tan \etal\cite{mnas} propose MNAS-Net, which performs NAS using an RL-agent for mobile devices.
Cai \etal\cite{proxyless_nas} propose ProxylessNAS, which derives specialized, hardware-specific CNN architectures from an over-parameterized model. 

Differently, Guo \etal\cite{dynamic} dynamically prune the CNN by learnable parameters, which have the ability to recover when required. Their scheme incorporates pruning in the training flow, resulting in irregular sparsity.
Zhang \etal\cite{structadmm} incorporate a cardinality constraint into the training objective to obtain different pruning regularity.
Bagherinezhad \etal\cite{lcnn} propose Lookup-CNN (LCNN), where the model learns a dictionary of shared filters at training time. During inference, the input is then convolved with the complete dictionary profiting from lower computational complexity. As the same accounts for filter-sharing, this weight-sharing approach is arguably closest to the technique developed in this work. However, the methodology itself and the training procedure are still fundamentally different from one another. 

{\renewcommand{\arraystretch}{1.2}
\begin{table}[h]
    \begin{center}
	\caption{Classification of model compression methods.}
	\vspace*{-2mm}
	\label{tab:compress_methods}
	    \resizebox{\columnwidth}{!}{
        \begin{tabular}{lccc}
        \toprule
        \multirow{2}{*}{\textbf{Method\textbackslash Advantage}} &\textbf{No Pre-trained} & \textbf{Learning} & \textbf{No Extensive}\\
		&  \textbf{Model} & \textbf{Policy} & \textbf{Exploration}\\
		\midrule
		\midrule
		\textbf{Rule-based Compression:}&&&\\
		\hspace{2mm}Low-Rank Dec.\cite{accelerating,tucker}& \xmark& \xmark & \xmark  \\
		\hspace{2mm}Prune (Handcrafted)\cite{learn_weights,dsc,geometric_mean_filter}& \xmark& \xmark & \xmark  \\
		
		\textbf{Learning-based Compression:}&&&\\
		\hspace{2mm}Prune (RL-Agent)\cite{learn_to_prune,amc}&  \xmark& \cmark & \xmark  \\
		\hspace{2mm}NAS\cite{mnas,proxyless_nas}&  \cmark& \cmark & \xmark  \\
		\hspace{2mm}Prune (Automatic)\cite{dynamic,lcnn, structadmm}&  \cmark& \cmark & \cmark \\
		\hspace{2mm}\textbf{ALF [Ours]} & \cmark & \cmark & \cmark \\
		\bottomrule
    	\end{tabular}}
        \setlength{\belowcaptionskip}{-12pt}
   \end{center}
   \vspace{-6mm}
\end{table}}

%% file: data/40_alf.tex
\begin{figure}[h!]
\centering
\includegraphics[width=0.94\columnwidth]{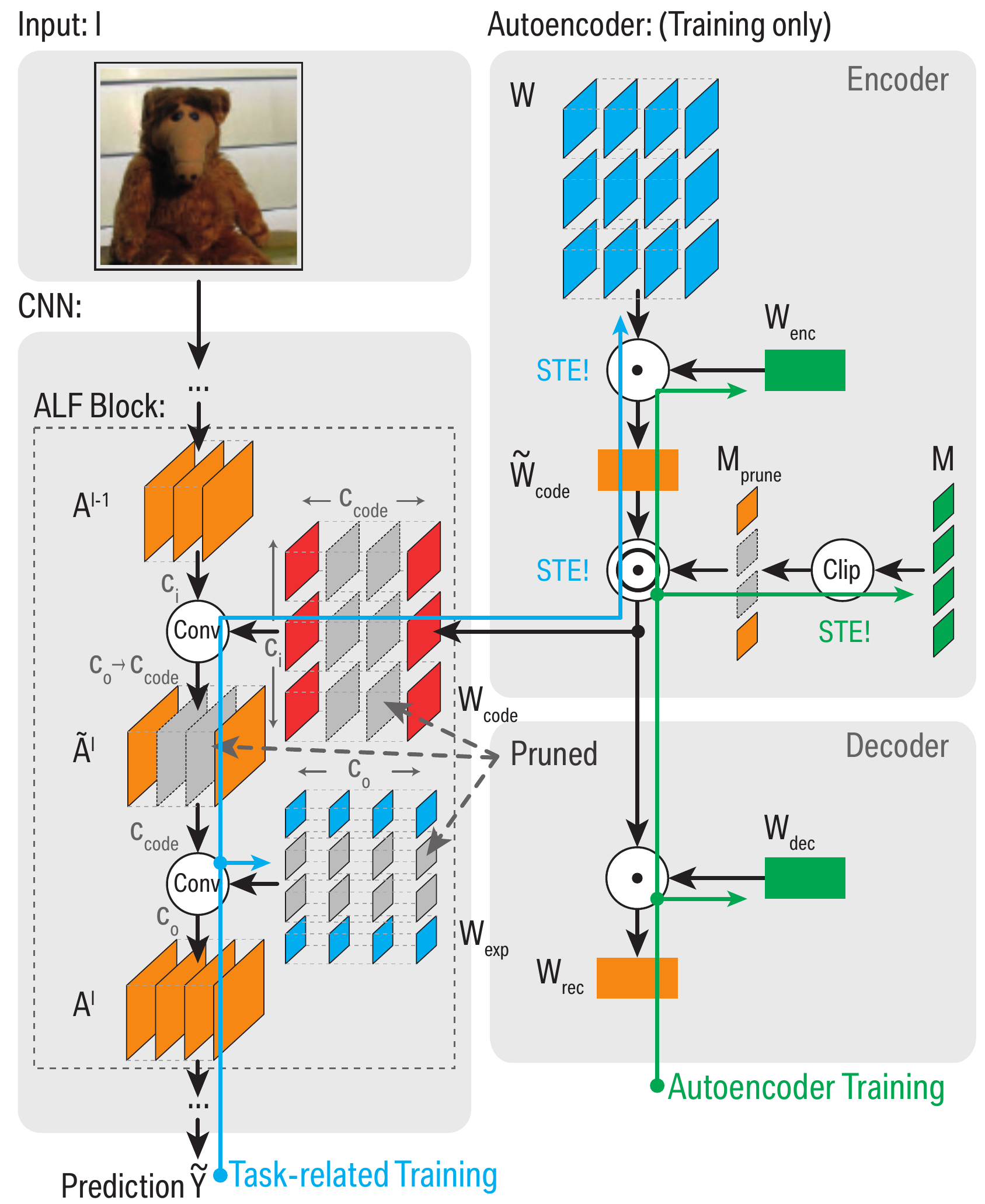}
\caption{ALF block in training mode \iffalse ($C_{\text{code}} = C_o$ )\fi showing the forward and backward path for the task-related and autoencoder training.}
\label{fig:alf_overview}
\vspace{-3ex}
\end{figure}

The goal of the proposed filter-sharing technique is to replace the standard convolution with a more efficient alternative, namely the \alf{} (ALF)-block. An example of the ALF-block is shown in Fig.~\ref{fig:alf_overview}.

Without loss of generality, \(A^{l-1} \in \mathbb{R}^{H_i \times W_i \times C_i}\) is considered as an input feature map to a convolutional layer \(l \in [1,L]\) of an $L$-layer CNN, where $H_i$ and $W_i$ indicate the height and width of the input, and $C_i$ is the number of input channels. The weights \(W^{l} \in \mathbb{R}^{K \times K \times C_i \times C_o}\) are the trainable parameters of the layer $l$, where $K$ and $C_o$ are the kernel dimensions and the number of output channels respectively. 

In detail, the task is to approximate the filter bank $W$ in a convolutional layer during training by a low-rank version \(W_{\text{code}} \in \mathbb{R}^{K \times K \times C_i \times C_{\text{code}}}\), where $C_{\text{code}}<C_o$. The low-rank version of the weights $W_{\text{code}}$ is utilized later in the deployment stage for an embedded-friendly application.

In contrast to previous structured pruning approaches \cite{lcnn,dsc,geometric_mean_filter}, this method does not intend to alter the structure of the model in a way which results in a changed dimensionality of the output feature maps \(A^{l} \in \mathbb{R}^{H_o \times W_o \times C_o}\), where $H_o$ and $W_o$ indicate the height and width of the output. This is done by introducing an additional expansion layer \cite{squeezenet}. The advantages are twofold. First, each layer can be trained individually without affecting the other layers. Second, it simplifies the end-to-end training and allows comparison of the learned features.

The expansion layer is comprised of point-wise convolutions with weights \(W_{\text{exp}} \in \mathbb{R}^{1 \times 1 \times C_{\text{code}} \times C_o}\), for mapping the intermediate feature maps after an ALF-block \(\tilde{A}^{l} \in \mathbb{R}^{H_o \times W_o \times C_{\text{code}}}\), to the output feature map $A^{l}$, as expressed in Eq.~\ref{eq:auto_share_inference}.
\begin{equation}
 \label{eq:auto_share_inference}
 A^{l} = \sigma(\tilde{A}^l \ast W_{\text{exp}}) = \sigma(\sigma_{\text{inter}}(A^{l-1} \ast W_{\text{code}}) \ast W_{\text{exp}})
\end{equation}
As the point-wise convolution introduces a certain overhead with regard to operations and weights, it is necessary to analyze the resource demands of the ALF-block compared to the standard convolution and ensure $C_{\text{code}}<C_{\text{code,max}}$, where $C_{\text{code,max}}$ denotes the number of filters which have to be removed to attain an efficiency improvement, see Eq.~\ref{eq:gain}.
\begin{equation}
\label{eq:gain}
\frac{C_i C_o K^2}{C_{\text{code}} (C_i K^2 + C_o)}\rightarrow C_{\text{code,max}}=\lfloor\frac{C_iC_oK^2}{C_iK^2+C_o}\rfloor
\end{equation}


\subsection{Autoencoder-based Low-rank Filter-sharing Block}
\input{data/41_alf_block.tex}

\subsection{Training Procedure}
\input{data/43_alf_training.tex}

\subsection{Deployment}
\input{data/44_alf_postprocessing.tex}

%% file: data/41_alf_block.tex
As stated before, the autoencoder is required to identify correlations in the original weights $W$ and to derive a compressed representation $W_{\text{code}}$ from them.
The autoencoder is only required in the training stage and is discarded in the deployment stage.

Referring back to Fig.~\ref{fig:alf_overview}, the autoencoder setup including the pruning mask $M_{\text{prune}} \in \mathbb{R}^{1 \times 1 \times 1 \times C_o}$ is illustrated.
According to the design of an autoencoder, Eq.~\ref{eq:code} gives the complete expression for calculating the compressed weights $W_{\text{code}}$. 
The encoder performs a matrix multiplication between the input $W$ and the encoder filters $W_{\text{enc}} \in \mathbb{R}^{K\times K\times C_o\times C_{\text{code}}}$. $M_{\text{prune}}$ zeroizes elements of $\tilde{W}_{\text{code}}$ and $\sigma_{\text{ae}}$ refers to a non-linear activation function, i.e. $\tanh()$. Different configurations of $\sigma_{\text{ae}}$, $\sigma_{\text{inter}}$ and initialization schemes are studied in Sec.~\ref{subsec:ablation}. 
\begin{equation}
\label{eq:code}
W_{\text{code}} = \sigma_{\text{ae}}(\tilde{W}_{\text{code}} \odot M_{\text{prune}})=\sigma_{\text{ae}}((W \cdot W_{\text{enc}}) \odot M_{\text{prune}})
\end{equation}

Eq.~\ref{eq:rec} provides the corresponding formula for the reconstructed filters $W_{\text{rec}}$ of the decoding stage. The symbol $\cdot$ stands for a matrix multiplication and $\odot$ for a Hadamard product respectively. 
The pruning mask $M_{\text{prune}}$ acts as a gate, allowing only the most salient filters to appear as non-zero values in $W_{\text{code}}$, in the same manner as sparse autoencoders. The decoder must, therefore, learn to compensate for the zeroized filters to recover a close approximate of the input filter bank.
\begin{equation}
\label{eq:rec}
W_{\text{rec}} = \sigma_{\text{ae}}(W_{\text{code}} \cdot W_{\text{dec}})
\end{equation}

In order to dynamically select the most salient filters, an additional trainable parameter, denoted mask \(M \in \mathbb{R}^{1 \times 1 \times 1 \times C_o}\), is introduced with its individual elements \(m_i \in M\). By exploiting the sparsity-inducing property of L1 regularization, individual values in the mask $M$ are driven towards zero during training. Since the optimizer usually reaches values close to zero, but not exactly zero, clipping is performed to zero out values that are below a certain threshold $t$. Further, the clipping function $M_{\text{prune}} = \clip(M, t) = \mathbb{I}_{\{\lvert m_i \lvert~> t\}}m_i$ allows the model to recover a channel when required. 

%% file: data/43_alf_training.tex
For understanding the training procedure of ALF, it is important to fully-comprehend the training setup, including the two player game of the CNN and the ALF-blocks.
\newline
\textbf{Task-related Training: }
The weights $W$ are automatically compressed by the ALF-blocks. Allowing the weights $W$ to remain trainable, instead of using fixed filters from a pre-trained model, is inspired by binary neural networks (BNNs)~\cite{bnn}. The motivation is that weights from a full-precision CNN might not be equally applicable when used in a BNN and, thus, require further training. Analogously, the weights from a pre-trained model might not be the best fit for the instances of $W$ in the filter-sharing use-case. Therefore, training these variables is also part of the task optimizer's job. It's objective is the minimization of the loss function $\mathcal{L}_{\text{task}}$, which is the accumulation of the cross-entropy loss $\mathcal{L}_{\text{CE}}$, of the model's prediction $\tilde{Y}$ and the label $Y$ of an input image $I$, and the weight decay scaling factor $\nu_{\text{wd}}$ multiplied with the L2 regularization loss $\mathcal{L}_{\text{reg}}$.

It is important to point out that no regularization is applied to the instances of either $W$ or $W_{\text{code}}$. Even though each $W_{\text{code}}$ contributes to a particular convolution operation with a considerable impact on the task loss $\mathcal{L}_{\text{task}}$, the task optimizer can influence this variable only indirectly by updating $W$. As neither of the variables $W_{\text{enc}}$, $W_{\text{dec}}$, or $M$ of the appended autoencoder are trained based on the task loss, they introduce a sufficiently high amount of noise, which arguably makes any further form of regularization more harmful than helpful.

In fact, this noise introduced by autoencoder variables does affect the gradient computation for updating \mbox{variable $W$}. As this might hamper the training progress, Straight-Through-Estimator (STE) \cite{bnn} is used as a substitute for the gradients of Hadamard product with the pruning mask, as well as for multiplication with the encoder filters. 
This ensures that the gradients for updating the input filters can propagate through the autoencoder without extraneous influence. 

This measure is especially important in case of the Hadamard product, as a significant amount of weights in $M_{\text{prune}}$ might be zero. When including this operation in the gradient computation, a correspondingly large proportion would be zeroized as a result, impeding the information flow in the backward pass. Nevertheless, this problem can be resolved by using the STE. 
In Eq.~\ref{eq:grad_w_orig}, the gradients for the variables $g_W$ for a particular ALF-block are derived.
\vspace{-3mm}
\begin{align}
\begin{split}\label{eq:grad_w_orig}
g_{W} &{}= \pdv{\mathcal{L}_{\text{task}}}{W} =\pdv{\mathcal{L}_{\text{task}}}{\tilde{A}} \cdot \pdv{\tilde{A}}{W_{\text{code}}} \cdot \pdv{W_{\text{code}}}{\tilde{W}_{\text{code}}} \cdot \pdv{\tilde{W}_{\text{code}}}{W_{\text{orig}}} \\
&\overset{\text{STE}}={} \pdv{\mathcal{L}_{\text{task}}}{\tilde{A}} \cdot \pdv{\tilde{A}}{W_{\text{code}}} 
\end{split}
\end{align}
\textbf{Autoencoder Training: }
Each autoencoder is trained individually by a dedicated SGD optimizer, referred to as an autoencoder optimizer. The optimization objective for such an optimizer lies in minimizing the loss function $\mathcal{L}_{\text{ae}}= \mathcal{L}_{\text{rec}} + \nu_{\text{prune}}\cdot \mathcal{L}_{\text{prune}}$. 
The reconstruction loss is computed using the MSE metric and can be expressed by $\mathcal{L}_{\text{rec}} = \mse(W, W_{\text{rec}})$. In the field of knowledge distillation \cite{CRD}, similar terms are studied.
The decoder must learn to compensate for the zeroized filters to recover a close approximate of the input filter. If a lot of values in $M_{\text{prune}}$ are zero, a large percentage of filters are pruned.
To mitigate this problem, the mask regularization function $\mathcal{L}_{\text{prune}} = 1/C_o \sum{|m|}$ is multiplied with a scaling factor $\nu_{\text{prune}}=\texttt{max}(0, 1-e^{(m*(\theta-{\text{pr}_{\text{max}}}))})$, which decays with increasing zero fraction in the mask and slows down the pruning rate towards the end of the training. In detail, $\nu_{\text{prune}}$ adopts the pruning sensitivity \cite{learn_weights} of convolutional layers, where $m\in[1,10]$ is the slope of the sensitivity, $pr_{\text{max}}\in[0,1]$ is the maximum pruning rate and $\theta=C_{\text{code,zero}}/C_{\text{code}}$ is the zero fraction, with $C_{\text{code,zero}}$ referring to the number of zero filters in $W_{\text{code}}$,
As a consequence, the regularization effect decreases and fewer filters are zeroized in the code eventually. In other words, the task of $\mathcal{L}_{\text{rec}}$ is to imitate $W$ by $W_{\text{rec}}$ by learning $W_{\text{dec}}$, $W_{\text{enc}}$ and $M$ while $\mathcal{L}_{\text{prune}}$ steadily tries to prune further channels.

The autoencoder optimizer updates the variables $W_{\text{enc}}$, $W_{\text{dec}}$ and $M$, based on the loss derived from a forward pass through the autoencoder network. Since $M_{\text{prune}}$ and $W_{\text{code}}$ are intermediate feature maps of the autoencoder, they are not updated by the optimizer. While the gradient calculation for the encoder and decoder weights is straight forward, the gradients for updating the mask $M$ require special handling. The main difficulty lies in the clipping function which is non-differentiable at the data points $t$ and $-t$. As previously mentioned, for such cases the STE can be used to approximate the gradients. 
The mathematical derivation for the gradients, corresponding to the variables updated by the autoencoder optimizer, are given in Eq.~\ref{eq:grad_m}.  
\begin{align}
\begin{split}\label{eq:grad_m}
g_{M} &{}= \pdv{\mathcal{L}_{\text{ae}}}{M} 
= \pdv{\mathcal{L}_{\text{ae}}}{W_{\text{rec}}} \cdot \pdv{{W_{\text{rec}}}}{W_{\text{code}}} \cdot \pdv{{W_{\text{code}}}}{M_{\text{prune}}} \cdot \pdv{{M_{\text{prune}}}}{M} \\
&\overset{\text{STE}}={} \pdv{\mathcal{L}_{ae}}{W_{\text{rec}}} \cdot \pdv{{W_{\text{rec}}}}{W_{\text{code}}} \cdot \pdv{{W_{\text{code}}}}{M_{\text{prune}}}
\end{split}
\end{align}

%% file: data/44_alf_postprocessing.tex
The utility of the autoencoders is limited to the training process, since the weights are fixed in the deployment stage. The actual number of filters is still the same as at the beginning of the training ($C_{code} = C_o$). However, the code comprises of a certain amount of filters containing only zero-valued weights which are removed. For the succeeding expansion layer, fewer input channels imply that the associated channels in $W_{\text{exp}}$ are not used anymore and can be removed as well (Fig.\ref{fig:alf_overview} pruned channels (gray)).
After post-processing, the model is densely compressed and is ready for efficient deployment. 


%% file: data/50_experiments.tex
We evaluate the proposed ALF technique on CIFAR-10~\cite{CIFAR_10} and ImageNet~\cite{imagenet_cvpr09} datasets. 
The 50k train and 10k test images of CIFAR-10 are used to respectively train and evaluate ALF. The images have a resolution of \(32\times32\) pixel. ImageNet consists of \(\sim1.28\) Mio. train and 50K validation images with a resolution of ($256\times256$) px. If not otherwise mentioned, all hyper-parameters specifying the task-related training were adopted from the CNN's base implementation. For ALF, the hyperparameters $m=8$ and $pr=0.85$ are set.

\subsection{Configuration Space Exploration} \label{subsec:ablation}
\input{data/51_experiments_ablation.tex}

\subsection{Comparison with State-of-the-Art}
\input{tab/hw_estimates_revised.tex}
\input{data/52_experiments_Cifar10.tex}
\input{data/52_experiments_HW_motivation.tex}

\input{data/52_experiments_ImageNet.tex}

%% file: data/51_experiments_ablation.tex
\begin{figure}
\begin{subfigure}{0.90\columnwidth}
  \centering
  \includegraphics[trim={0 2.1cm 2.66cm 2.1cm},width=0.95\columnwidth]{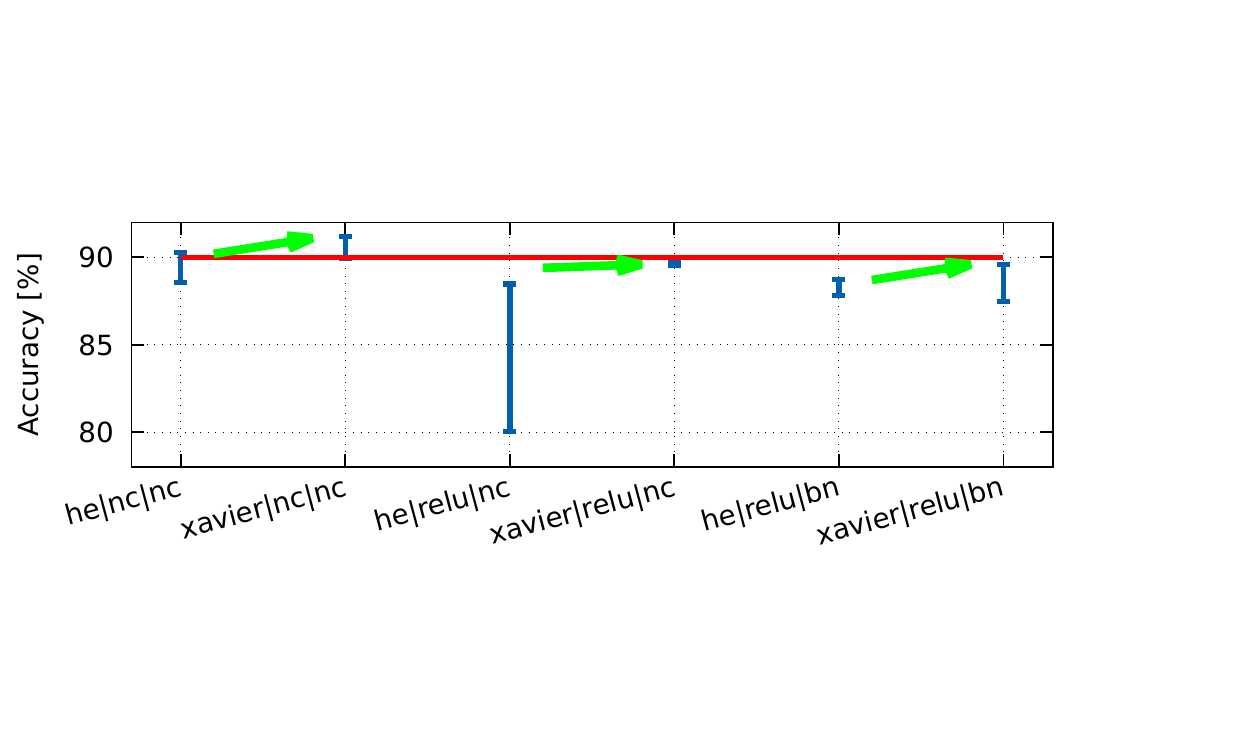}
  \caption{Configuration [$W_{\text{exp,init}}|\sigma_{\text{inter}}|\text{BN}_{\text{inter}}$].}
  \label{fig:setup1}
\end{subfigure}%
\\
\begin{subfigure}{0.90\columnwidth}
  \centering
  \includegraphics[trim={0 2.1cm 2.4cm 1.9cm},width=0.95\columnwidth]{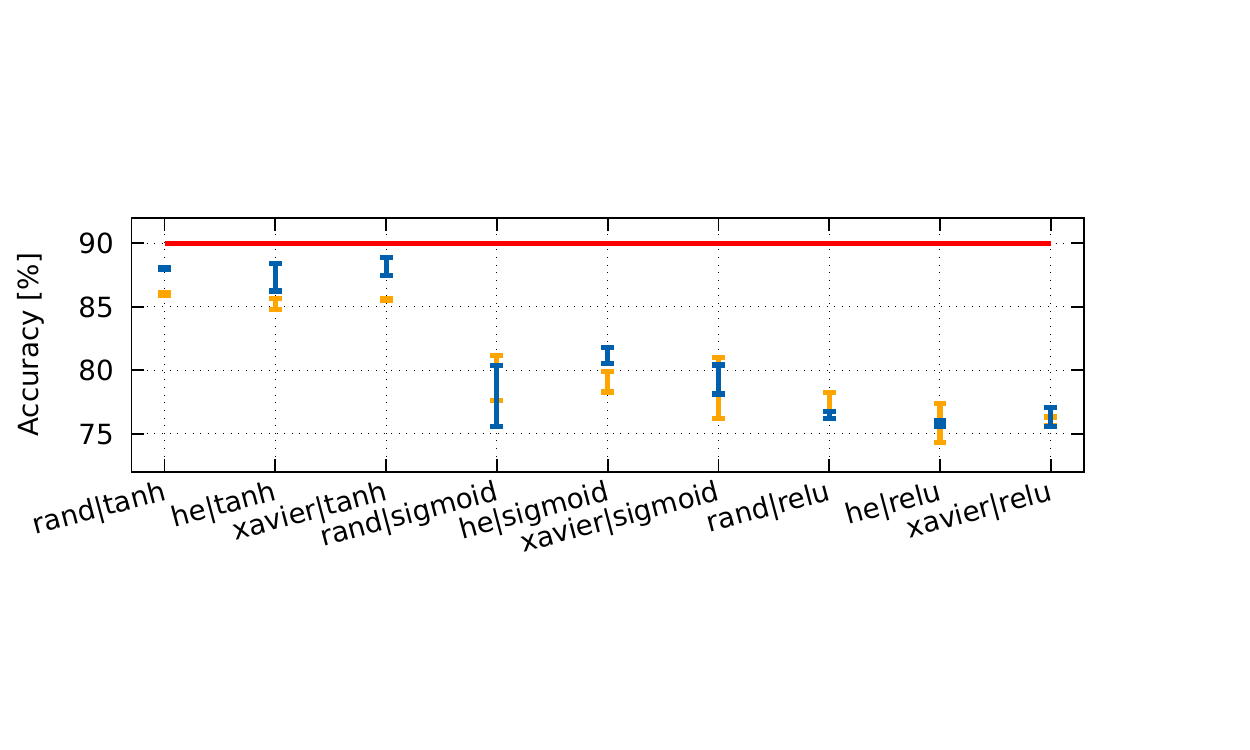}
  \caption{Config. [$W_{\text{ae,init}}|\sigma_{\text{ae}}$], $\sigma_{\text{inter}}$=\textcolor{blue}{none}), $\sigma_{\text{inter}}$=\textcolor{orange}{ReLU}.}
  \label{fig:sfig2}
\end{subfigure}
\begin{subfigure}{0.85\columnwidth}
  \centering
  \includegraphics[trim={1cm 1.9cm 2.5cm 1.6cm},width=0.85\columnwidth]{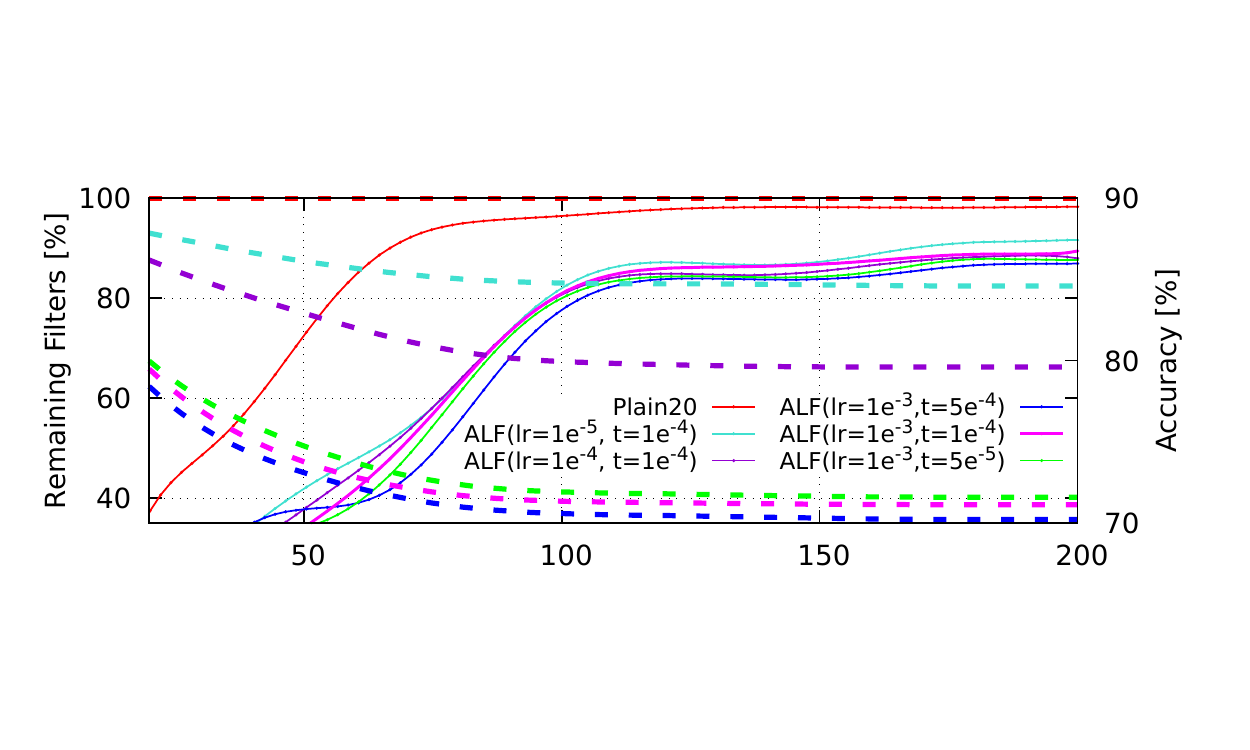}
  \caption{Training epochs. ALF trained on different $lr$ and $t$.}
  \label{fig:sfig3}
\end{subfigure}
\caption{Design space exploration for Plain-20 on CIFAR-10.}
\label{fig:fig_base_model_init}
\vspace{-5mm}
\end{figure}

Crucial design decisions are investigated in this section. 
The aforementioned novel training setup includes a number of new parameters, namely the weight initialization scheme for $W_{\text{exp}}, W_{\text{enc}}$ and $W_{\text{dec}}$, consecutive activation functions and batch normalization layers.
For that purpose, Plain-20~\cite{resnet} is trained on CIFAR-10. Experiments are repeated at least twice to provide results with decent validity (bar-stretching).
\newline
\noindent\textbf{Setup 1:} The effect of additional expansion layers is studied in Fig.~\ref{fig:setup1} taking the initialization (He\cite{he} and Xavier \cite{xavier}), extra non-linear activations $\sigma_{\text{inter}}$ (i.e. ReLU) and batch normalization (BN) into account.
The results suggest that expansion layers can lead to a tangible increase in accuracy. In general, the Xavier initialization yields slightly better results than the He initialization and is chosen for the expansion layer of the ALF blocks. In addition, the incorporation of the $\text{BN}_{\text{inter}}$ layer seems to not have perceivable advantages. The influence of $\sigma_{\text{inter}}$ is confirmed in the next experiment.
\newline
\textbf{Setup 2:} In this setup, the ALF block's pruning mask $M$ is not applied, thus, no filters are pruned. This is applicable to select a weight initialization scheme for $W_{\text{enc}}$ and $W_{\text{dec}}$ (referred as $W_{\text{ae,init}}$) and an activation function $\sigma_{\text{ae}}$. In case no activation function $\sigma_{\text{inter}}$ (\textcolor{blue}{blue}) is applied to $\tilde{A}^l$, the accuracy is higher than with ReLU layers. Based on the results the Xavier initialization is selected for $W_{ae,init}$. Moreover, $\texttt{tanh}()$ outperforms other non-linear activation functions $\sigma_{ae}$. As the pruning mask $M$ is not active, the regularization property of the autoencoder is absent causing a noticeable accuracy drop. 
\newline
\textbf{Setup 3:}
Five variants of ALF, resulting in different sparsity rates, are explored in Fig~\ref{fig:sfig3} and compared to the uncompressed Plain-20 (90.5\% accuracy).
The first three variants differ in terms of the threshold $t \in\{5\cdot10^{-5},1\cdot10^{-4},5\cdot10^{-4}\}$ while the learning rate $lr_{\text{ae}}=1\cdot10^{-3}$. We observe that the pruning gets more aggressive when the threshold $t$ is increased. The number of non-zero filters remaining are $40.17\%$, $38.6\%$ and $35.71\%$ respectively (see green, pink and blue curves).  
We select the threshold $t=1e^{-4}$ as a trade-off choice between sparsity and accuracy. The behaviour of ALF is also accessed by changing the learning rate of the autoencoder. The learning rate $lr_{\text{ae}}\in\{1\cdot10^{-5}, 1\cdot10^{-4}\}$ is explored in the consecutive variants (see turquoise and purple curves). The number of remaining non-zero filters increases as there are less updates to the sparsity mask $M$. In case of $lr_{\text{ae}}= 1\cdot10^{-4}$ (purple), the resulting network has less number of non-zero filters with a significant accuracy drop. However, considering the trade-off between accuracy and pruning rate, we choose the learning rate $lr_{\text{ae}}=1\cdot10^{-3}$ for the autoencoder.  

%% file: tab/hw_estimates_revised.tex
\begin{figure*}[h!]
    \centering
    \resizebox{\linewidth}{!}{
		\begin{tikzpicture}
		\begin{axis}[
		name=vanilla,
		height=11cm,
		width=60cm,
		ybar stacked,
		ymajorgrids=true,
		bar shift=24pt,
        legend image post style={scale=1.7},
        legend style={at={(0.99,0.92)},anchor=east, legend columns=-1, draw = none, nodes={scale=1.5, transform shape}, column sep=5pt},
		ylabel={Normalized Energy},
		symbolic x coords={CONV1, CONV211, CONV212, CONV221, CONV222, CONV231, CONV232, CONV311, CONV312, CONV321, CONV322, CONV331, CONV332, CONV411, CONV412, CONV421, CONV422, CONV431, CONV432},
		xtick=data,
		xticklabel style = {font=\Large, yshift=-2ex},
		yticklabel style = {font=\Large},
		ylabel style = {font=\huge, yshift=3ex},
		ymin=0,
        ymax=400000000,
        enlarge x limits={0.05},
		cycle list name=colorlist1,
		]
		\addplot coordinates {(CONV1, 28686506) (CONV211,151966868) (CONV212,151966868) (CONV221,151966868) (CONV222,151966868) (CONV231,151966868) (CONV232,151966868) (CONV311,76373539) (CONV312,152155669) (CONV321,152155669) (CONV322,152155669) (CONV331,152155669) (CONV332,152155669) (CONV411,76534149) (CONV412,152396650) (CONV421,152396650) (CONV422,152396650) (CONV431,152396650) (CONV432,152396650)};
		
		\addplot coordinates {(CONV1,7153892) (CONV211,14084980) (CONV212,14084980) (CONV221,14084980) (CONV222,14084980) (CONV231,14084980) (CONV232,14084980) (CONV311,7859752) (CONV312,9526513) (CONV321,9526513) (CONV322,9526513) (CONV331,9526513) (CONV332,9526513) (CONV411,6833908) (CONV412,6856662) (CONV421,6856662) (CONV422,6856662) (CONV431,6856662) (CONV432,6856662)};
		
		\addplot coordinates {(CONV1,63612800) (CONV211,112076800) (CONV212,112076800) (CONV221,112076800) (CONV222,112076800) (CONV231,112076800) (CONV232,112076800) (CONV311,82892800) (CONV312,77824000) (CONV321,77824000) (CONV322,77824000) (CONV331,77824000) (CONV332,77824000) (CONV411,51814400) (CONV412,52428800) (CONV421,52428800) (CONV422,52428800) (CONV431,52428800) (CONV432,52428800)};
		\legend{Register, Global Buffer, Off-Chip DRAM}
		\node[align=center, font=\large, rotate=90] (N1) at (28,300) {Plain-20 / ResNet20};
		\end{axis}

		\begin{axis}[
		name=alfplain20cifarAlex,
		width=60cm,
		height=11cm,
		ybar stacked,
		bar shift=0pt,
        legend style={legend pos=north east, legend columns=-1, draw = none},
        symbolic x coords={CONV1, CONV211, CONV212, CONV221, CONV222, CONV231, CONV232, CONV311, CONV312, CONV321, CONV322, CONV331, CONV332, CONV411, CONV412, CONV421, CONV422, CONV431, CONV432},
		ticks=none,
		ymin=0,
        ymax=400000000,
        enlarge x limits={0.05},
		cycle list name=colorlist1,
		postaction={pattern=horizontal lines, very thick,pattern color=white},
		]
		\addplot coordinates {(CONV1, 28686506) (CONV211,105986328) (CONV212,140529443) (CONV221,105986328) (CONV222,105986328) (CONV231,95494825) (CONV232,85003321) (CONV311,26501153) (CONV312,99017521) (CONV321,74257672) (CONV322,47854011) (CONV331,21579148) (CONV332,21579148) (CONV411,20556135) (CONV412,35225845) (CONV421,37155227) (CONV422,39846460) (CONV431,37155227) (CONV432,21323505)};
		
		\addplot coordinates {(CONV1,7153892) (CONV211,19955617) (CONV212,21753216) (CONV221,19955617) (CONV222,19955617) (CONV231,19356418) (CONV232,18757218) (CONV311,9069528) (CONV312,94150170) (CONV321,11635088) (CONV322,8950067) (CONV331,7964042) (CONV332,7964042) (CONV411,4918744) (CONV412,4778424) (CONV421,4854273) (CONV422,4930121) (CONV431,4854273) (CONV432,4399184)};
		
		\addplot coordinates {(CONV1,63612800) (CONV211,177472000) (CONV212,197228800) (CONV221,177472000) (CONV222,177472000) (CONV231,170886400) (CONV232,164300800) (CONV311,97033600) (CONV312,91596800) (CONV321,84032000) (CONV322,75232000) (CONV331,66432000) (CONV332,66432000) (CONV411,49420800) (CONV412,40576000) (CONV421,41113600) (CONV422,41651200) (CONV431,41113600) (CONV432,37888000)};
		\node[align=center, font=\large, rotate=90] (N1) at (-3,300) {ALF-Plain-20};
		\end{axis}
		
		\begin{axis}[
		name=alfresnet20cifar,
		width=60cm,
		height=11cm,
		ybar stacked,
		bar shift=-24pt,
        legend style={legend pos=north east, legend columns=-1, draw = none},
        symbolic x coords={CONV1, CONV211, CONV212, CONV221, CONV222, CONV231, CONV232, CONV311, CONV312, CONV321, CONV322, CONV331, CONV332, CONV411, CONV412, CONV421, CONV422, CONV431, CONV432},
		ticks=none,
		ymin=0,
        ymax=400000000,
        enlarge x limits={0.05},
		cycle list name=colorlist1,
		postaction={pattern=north east lines, very thick,pattern color=white},
		]
		\addplot coordinates {(CONV1, 28686506) (CONV211,105986328) (CONV212,105986328) (CONV221,95494825) (CONV222,85003321) (CONV231,74511816) (CONV232,95494825) (CONV311,26501153) (CONV312,70404851) (CONV321,74257672) (CONV322,42596361) (CONV331,21579148) (CONV332,21579148) (CONV411,19809832) (CONV412,29208513) (CONV421,37155227) (CONV422,29208513) (CONV431,31833451) (CONV432,29208513)};
		
		\addplot coordinates {(CONV1,7153892) (CONV211,19955617) (CONV212,19955617) (CONV221,19356418) (CONV222,18757218) (CONV231,18158019) (CONV232,19356418) (CONV311,9069528) (CONV312,11316526) (CONV321,11635088) (CONV322,8752862) (CONV331,7964042) (CONV332,7964042) (CONV411,4842895) (CONV412,4626728) (CONV421,4854273) (CONV422,4626728) (CONV431,4702577) (CONV432,4626728)};
		
		\addplot coordinates {(CONV1,63612800) (CONV211,177472000) (CONV212,177472000) (CONV221,170886400) (CONV222,164300800) (CONV231,157715200) (CONV232,170886400) (CONV311,97033600) (CONV312,82272000) (CONV321,84032000) (CONV322,73472000) (CONV331,66432000) (CONV332,66432000) (CONV411,48940800) (CONV412,39500800) (CONV421,41113600) (CONV422,39500800) (CONV431,40038400) (CONV432,39500800)};
		\node[align=center, font=\large, rotate=90] (N1) at (-32,300) {ALF-ResNet-20};
		\end{axis}
		
		\begin{axis}[
		name=latencyall,
		width=60cm,
		height=11cm,
        legend style={at={(0.99,0.85)},anchor=east, legend columns=-1, draw = none, nodes={scale=1.5, transform shape}, column sep=5pt},
		ylabel={Normalized Latency},
        symbolic x coords={CONV1, CONV211, CONV212, CONV221, CONV222, CONV231, CONV232, CONV311, CONV312, CONV321, CONV322, CONV331, CONV332, CONV411, CONV412, CONV421, CONV422, CONV431, CONV432},
		ylabel near ticks, yticklabel pos=right,
        yticklabel style = {font=\Large},
		ylabel style = {font=\huge, yshift=-3ex},
		xticklabels=\empty,
		ymin=0,
        ymax=2000000,
        enlarge x limits={0.05},
		cycle list name=colorlist2,
		]

				\addplot [mark=x, mark options={scale=5}, draw= Green3,line width=5pt, shift={(-28.5,0)}] coordinates {(CONV1,655361) (CONV211,1245184) (CONV212,1245184) (CONV221,1122305) (CONV222,999424) (CONV231,876544) (CONV232,1122305) (CONV311,95681) (CONV312,212480) (CONV321,226304) (CONV322,208896) (CONV331,93184) (CONV332,93184) (CONV411,104576) (CONV412,97792) (CONV421,111360) (CONV422,97792) (CONV431,133120) (CONV432,97792)};
		
		
        \addplot [mark=x, mark options={scale=5}, draw= Green2,line width=5pt, shift={(0,0)}] coordinates {(CONV1,655361) (CONV211,1245184) (CONV212,1476608) (CONV221,1245184) (CONV222,1245184) (CONV231,1122305) (CONV232,999424) (CONV311,95681) (CONV312,1258881) (CONV321,226304) (CONV322,103169) (CONV331,93184) (CONV332,93184) (CONV411,111360) (CONV412,104576) (CONV421,111360) (CONV422,118145) (CONV431,111360) (CONV432,90112)};

				\addplot [mark=x, mark options={scale=5}, draw= Green4, line width=5pt, shift={(28.5,0)}] coordinates {(CONV1, 655361) (CONV211,1703936) (CONV212,1703936) (CONV221,1703936) (CONV222,1703936) (CONV231,1703936) (CONV232,1703936) (CONV311,425984) (CONV312,425984) (CONV321,425984) (CONV322,425984) (CONV331,425984) (CONV332,425984) (CONV411,212992) (CONV412,212992) (CONV421,212992) (CONV422,212992) (CONV431,212992) (CONV432,212992)};
		
		\end{axis}
		
		\begin{axis}[
		name=legend,
		width=60cm,
		height=11cm,
		hide axis,
        legend style={at={(0.99,0.82)},anchor=east, legend columns=-1, draw = none, nodes={scale=1.5, transform shape}, column sep=5pt},
		ymin=0,
        ymax=1,
        xmin=0,
        xmax=1,
		cycle list name=colorlist2,
		]
		\addlegendimage{mark=x, mark options={scale=5}, draw= Green3, line width=5pt}
		\addlegendentry{ALF-ResNet-20};
		\addlegendimage{mark=x, mark options={scale=5}, draw= Green2, line width=5pt}
		\addlegendentry{ALF-Plain-20};
		\addlegendimage{mark=x, mark options={scale=5}, draw= Green4, line width=5pt}
		\addlegendentry{Plain-20/ResNet-20};
		\end{axis}
		\end{tikzpicture}}
		\vspace{-3ex}
		\caption{Energy consumption breakdown and latency of vanilla and ALF-compressed ResNet/Plain-20 execution.}
		\vspace{-5mm}
	\label{fig:mem_energy_lat}
\end{figure*}
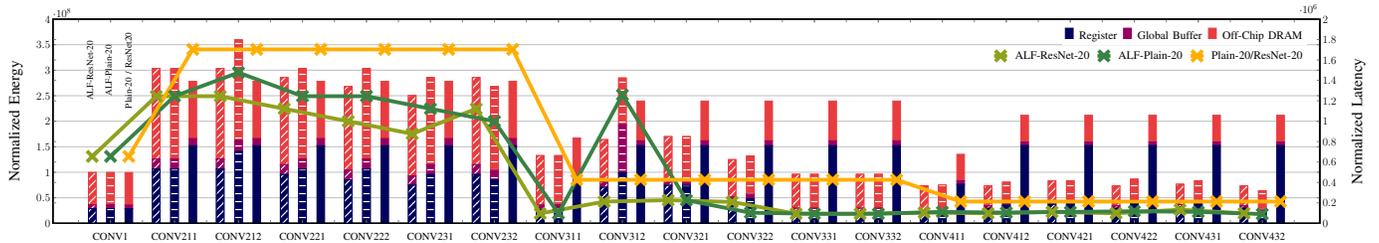

%% file: data/52_experiments_Cifar10.tex
\noindent\textbf{CIFAR-10:} Tab.~\ref{tab:benchmarking_cifar} compares ALF against its full precision counterpart (ResNet-20) and state-of-the-art pruned models \cite{amc,geometric_mean_filter}.
The ALF model consists of the least number of operations and training parameters compared to other pruning works, with an accuracy drop of 1.9\% compared to its full precision model.
\input{tab/tab_alf_cifar_bechmark.tex}

%% file: tab/tab_alf_cifar_bechmark.tex
\begin{table}[!h]
\centering
\caption[Benchmarking results on Cifar-10.]{Pruned CNNs on CIFAR-10, for Conv layers only.}
  \label{tab:benchmarking_cifar}
  \resizebox{\columnwidth}{!}{
  \begin{tabular}{llccc}
  \toprule
  \textbf{Method}&\textbf{Policy}&$\mathtt{Params}$&$\mathtt{OPs}$[$10^6$]&\textbf{Acc}\textbf{[\%]}
  \\
  \midrule
  \midrule
  Plain-20\cite{resnet}   & --- & 0.27M & 81.1 & 90.5
  \\
  ResNet-20\cite{resnet}  & --- & 0.27M & 81.1 & 91.3
  \\
  \midrule
  AMC~\cite{amc} & RL-Agent &  0.12M (-55\%) & 39.4 (-51\%) & 90.2
  \\
  FPGM~\cite{geometric_mean_filter} & Handcrafted & --- & 36.2 (-54\%) & \textbf{90.6}
  \\
    ALF (ours) & \multirow{2}{*}{Automatic} & \multirow{2}{*}{\textbf{0.07M (-70\%)}} & \multirow{2}{*}{\textbf{31.5 (-61\%)}} & \multirow{2}{*}{89.4}  
    \\
    ($t=10^{-4}$) && & 
    \\
    \bottomrule
    \end{tabular}}
    \vspace{-6mm}
\end{table}

%% file: data/52_experiments_HW_motivation.tex

\noindent\textbf{Hardware-Model Estimates:} Improvements in the number of $\mathtt{OPs}$ and $\mathtt{Params}$ of a CNN do not always indicate tangible advantages on real hardware execution. Here, we analyze the real advantage of ALF by investigating its implications on hardware. The deterministic execution of CNNs on target hardware platforms eases the creation of hardware models with highly accurate estimates of normalized latency and energy. 
We use an accurate modeling framework, Timeloop~\cite{timeloop}, to replicate the execution and scheduling scheme of the Eyeriss~\cite{EyerissSpatial} accelerator. The Eyeriss model used in the experiments consists of a $16\times16$ array of processing elements (PEs). Each PE contains three separate register files (RFs), one for each datatype (inputs, weights and outputs). The word-width of all datatypes is fixed to 16-bits. The combined RFs in a single PE add up to 220 words. The global buffer has a total size of 128KB and can hold output and input datatypes. Weights bypass the global buffer and are directly fed to the weight RFs of the PEs. The mapper's search algorithm follows an exhaustive method with a timeout of 100K iterations and victory condition of 1K iterations per thread. 
We simulate the compressed configurations achieved by the ALF-block on the Plain-20 and ResNet-20 models. The batch size for all experiments is set to 16 and the energy values are normalized against the energy cost of a single register file read \cite{EyerissSpatial}. The latency is normalized to the bandwidth of a register \mbox{(2 bytes/cycle)}.

Fig.~\ref{fig:mem_energy_lat} shows the advantages of the ALF technique.
The results show a trend of high contributions from the RF memory, particularly in the deeper layers of each CNN. This is due to the efficient dataflow of the accelerator, which increases the data reuse at the lowest-level memory, reducing the number of redundant data accesses between the higher memory levels.

By analyzing the energy breakdown results of both ALF models, an increase in DRAM energy is observed. This is due to additional off-chip data movement introduced by the expansion layer.
This is highlighted in the initial layers, as their inputs are larger. Practically, such codependent layers can be fused with some advanced scheduling techniques~\cite{layer_fusion}, eliminating this overhead. Nevertheless, the strong improvements in the deeper layers offset this degradation, leading to an overall 29\% lower energy consumption and 41\% latency reduction over the vanilla model.

The latency line-plots reveal an anomaly for layer $\texttt{conv312}$ on the ALF-Plain-20 compressed model. This compressed layer requires more energy and execution cycles than the vanilla Plain-20 execution. Taking a closer look, ALF-Plain-20 utilizes only 8\% of the PEs for processing this layer. 
Although ALF is a structured pruning technique, the resulting non-sparse model could have reduced parallelism opportunities under the restrictions enforced by the row-stationary dataflow on the hardware. 
These results cannot be intuitively concluded by simply considering $\mathtt{OPs}$ and $\mathtt{Params}$ as metrics for a compression technique. This emphasizes the importance of hardware-aware validation of pruning advantages, as performed in this subsection.

%% file: data/52_experiments_ImageNet.tex
\input{tab/tab_ilsvrc_revised.tex}

\noindent\textbf{ImageNet}: 
Tab.~\ref{tab:benchmarking_ilsvrc_alf} compares ALF against other pruned models, i.e. LCNN\cite{lcnn} and FPGM\cite{geometric_mean_filter}. For comparison, GoogleNet~\cite{googlenet}, SqueezeNet~\cite{squeezenet} and ResNet-18~\cite{resnet} are also detailed in the table.
Compared to SqueezeNet, GoogleNet and ResNet, ALF requires $\times1.4$, $\times2.4$, $\times3.0$ less $\mathtt{OPs}$, respectively.
Moreover, ALF outperforms GoogleNet and ResNet w.r.t. the parameters.
Compared to other pruned models, ALF lies on the pareto-front for the number of parameters, the operations and the accuracy. On one hand, LCNN obtained a very competitive computational complexity ($-40\%$), at the cost of an accuracy degradation of $-2.1\%$ compared to ALF.  On the other hand, AMC and FPGM gain $+3.4\%$, $+3.5\%$ accuracy improvement, but has $+51\%$, $+76\%$, more $\mathtt{OPs}$ than ALF respectively.

%% file: tab/tab_ilsvrc_revised.tex
\begin{table}[!h]
\vspace{-1ex}
\centering
\caption[Benchmarking results on ImageNet.]{Benchmarking results on ImageNet, comparing models and methods with regard to $\mathtt{MACs}$ and $\mathtt{Params}$.}
  \label{tab:benchmarking_ilsvrc_alf}
  \resizebox{\columnwidth}{!}{
  \begin{tabular}{llccc}
  \toprule
  \textbf{Method}&\textbf{Policy}&$\mathtt{Params}$&$\mathtt{OPs}$ [$10^6$]&\textbf{Acc [\%]}\\
    \midrule
    \midrule
    SqueezeNet~\cite{squeezenet}        & ---       & \textbf{1.23M}& 1722 & 57.2~\% \\ 
    GoogleNet~\cite{googlenet}          & ---       & 6.80M          & 3004 & 66.8\%\\
    ResNet-18~\cite{resnet}             & ---       & 11.83M        & 3743& \textbf{69.8~\%} \vspace{1mm}\\ 
    \multicolumn{2}{l}{\textbf{Pruned ResNet-18:}}&&&\\
    \midrule
    LCNN~\cite{lcnn}                    & Automatic    & --     & \textbf{749} & 62.2~\% \\
    FPGM~\cite{geometric_mean_filter}   & Handcrafted& --     & 2178         & 67.8~\% \\ 
    AMC~\cite{amc}                      & RL-Agent  & 8.9M    & 1874         & 67.7~\%\\
    ALF (ours)           & Automatic ($t=10^{-4}$)   & 4.24M    & 1239          & 64.3~\%\\
    \bottomrule
    \end{tabular}}
    \vspace{-1ex}
\end{table}

%% file: data/60_conclusion.tex
Pruning is a promising compression technique, ranging from rule-based to handcrafted and learning-based methods. 
Most approaches require either a pre-trained model or an extensive model exploration, making the compression a time consuming process.
In this paper, we propose the \alf{} technique (ALF) to dynamically prune a given CNN during task-specific training. ALF employs a sparse autoencoder to approximate the weight filters of the CNN for an embedded-friendly application. The novel method is applied to computer vision tasks, CIFAR-10 and ImageNet. Comparison to state-of-the-art methods is performed on well known theoretical metrics ($\mathtt{OPs}$ and $\mathtt{Params}$), as well as an analysis on hardware-model estimates, i.e. normalized latency and energy.